# Modelling of automotive steel fatigue lifetime by machine learning method

Oleh Yasniy [1,†], Dmytro Tymoshchuk [1,*,†], Iryna Didych [1,†], Nataliya Zagorodna [1,†] and Olha Malyshevska [2,†]

[1] *Ternopil Ivan Puluj National Technical University, Ruska str. 56, Ternopil, 46001, Ukraine*
[2] *Ivano-Frankivsk National Medical University, Galytska Str. 2, Ivano-Frankivsk, 76018, Ukraine*

**Abstract**
In the current study, the fatigue life of QSTE340TM steel was modelled using a machine learning method, namely, a neural network. This problem was solved by a Multi-Layer Perceptron (MLP) neural network with a 3-75-1 architecture, which allows the prediction of the crack length based on the number of load cycles $N$, the stress ratio $R$, and the overload ratio $Rol$. The proposed model showed high accuracy, with mean absolute percentage error (MAPE) ranging from 0.02% to 4.59% for different $R$ and $Rol$. The neural network effectively reveals the nonlinear relationships between input parameters and fatigue crack growth, providing reliable predictions for different loading conditions.

**Keywords**
machine learning, neural network, fatigue life, crack length, QSTE340TM steel

## 1. Introduction

QSTE340TM steel is a thermomechanically hardened low-alloy steel used in the automotive and mechanical engineering industries. Due to its high strength, QSTE340TM steel can reduce the weight of structures, which is important for automotive parts such as chassis, suspensions, and body components. It has good fatigue resistance, which ensures durability in harsh environments. The chemical composition of the steel includes manganese, silicon, phosphorus, sulfur, and other alloying elements that give it the required mechanical properties [1].

Machine learning methods allow us to model the fatigue life of QSTE340TM steel and effectively predict the material's durability under cyclic loading. By applying machine learning algorithms, a large amount of experimental data can be analyzed automatically and the relationship between various parameters affecting material properties can be determined [2,3,4,5,6].

## 2. Methods

Neural networks allow us to model the fatigue life of QSTE340TM steel and effectively predict crack growth in the material under cyclic loading. Functional dependencies were modelled for experimental data obtained in [7]. The dataset [8] contained the dependence of the crack length $a$ on the number of loading cycles $N$ for four stress ratios $R$, namely, $R$ = 0.1, 0.3, 0.5, and 0.7 at a constant amplitude (CA) and after a single tensile overload with overload ratios $Rol$ = 1.5, 2.0. The neural network was trained on a dataset where the input parameters are the number of loading cycles $N$, the stress ratio $R$, and the overload ratio $Rol$, and the output parameter is the crack length $a$. The load cycle $N$ reflects the number of cycles the steel has been loaded and is one of the main parameters for assessing fatigue crack growth. The stress ratio $R$ determines the ratio of the minimum and maximum loads of the cycle, which also affects the rate of fatigue crack development. The overload ratio $Rol$ considers cases where the load exceeds the nominal values.

The first 80% of the load cycles were used for the training, testing, and validation process. The accuracy of crack length prediction as a function of $N$, $R$, and $Rol$ was tested on the data of the next 20% loading cycles. During the training process, the dataset was divided into 3 parts: training, testing, and verification. The training, testing, and validation samples contained 1791 items, 80% of which were randomly selected for the training sample, 10% for the validation sample, and 10% for testing and evaluating the model's prediction quality. The forecasting error was calculated using the formula for the mean absolute percentage error (MAPE):

$$MAPE = 100\% \cdot \frac{1}{n} \sum_{i=1}^{n} \frac{\left| a_{true}^{test}(i) - a_{pred.}^{test}(i) \right|}{\left| a_{true}^{test}(i) \right|}, \quad (1)$$

where $n$ is the size of the test dataset, $a_{true}^{test}(i)$ is the true value of the crack length in the test dataset, $a_{pred}^{test}(i)$ is the predicted value of the crack length in the test dataset.

## 3. Results and discussion

The Multi-Layer Perceptron (MLP) neural network [9] 3-75-1 was used to predict the crack length $a$ in QSTE340TM steel depending on the number of load cycles $N$, the stress ratio $R$, and the overload ratio $Rol$. An effective model was created to detect nonlinear dependencies between these parameters accurately. The network consists of three layers: input, hidden, and output. The input layer contains three nodes corresponding to the three main input parameters. The hidden layer, which consists of 75 neurons, has the Tangential activation function. The output layer contains a single neuron designed to predict the crack length a, using a linear activation function (Linear), which allows the generation of continuous values without limiting their range, which is critical for adequately reflecting the actual processes of fatigue crack growth. This architecture allows the MLP 3-75-1 neural network to learn from data and accurately predict fatigue crack growth in QSTE340TM steel. Figure 1 shows the relationship between the experimental crack length values $a_{true}$ and the predicted values $a_{pred}$ obtained by the neural network for the test data set.

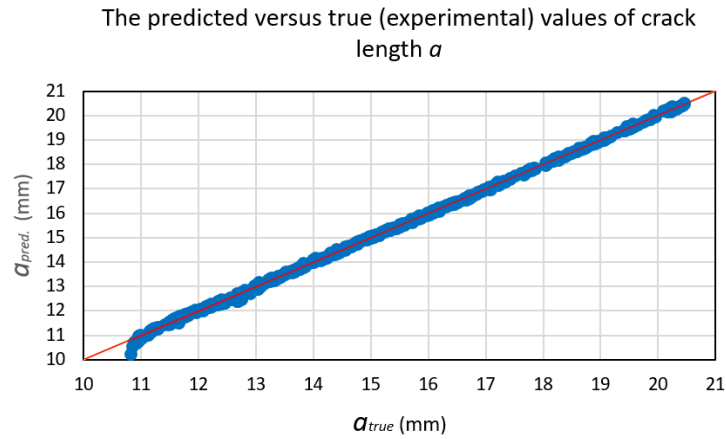

**Figure 1:** The predicted versus true (experimental) values of crack length $a$

As can be seen from Figure 1, the model is highly accurate, as almost all points lie along the bisector of the first coordinate angle, which means that the predicted values are almost identical to the experimental ones. The prediction error calculated by MAPE is only 0.34%. This indicates that the neural network effectively models fatigue crack growth in QSTE340TM steel, providing accurate predictions with minimal deviation from the actual values.

To test the accuracy of crack length prediction as a function of $N$, $R$, and $R_{ol}$, data from the next 20% of loading cycles were used. This data was removed at the initial stage and was not used to test the model accuracy. Figure 2 shows the relationship between the experimental crack length values $a_{true}$ and the predicted values $a_{pred}$, obtained for a stress ratio $R = 0.1$ at constant amplitude and with a single overload with overload factors $Rol$ = 1.5, 2.0.

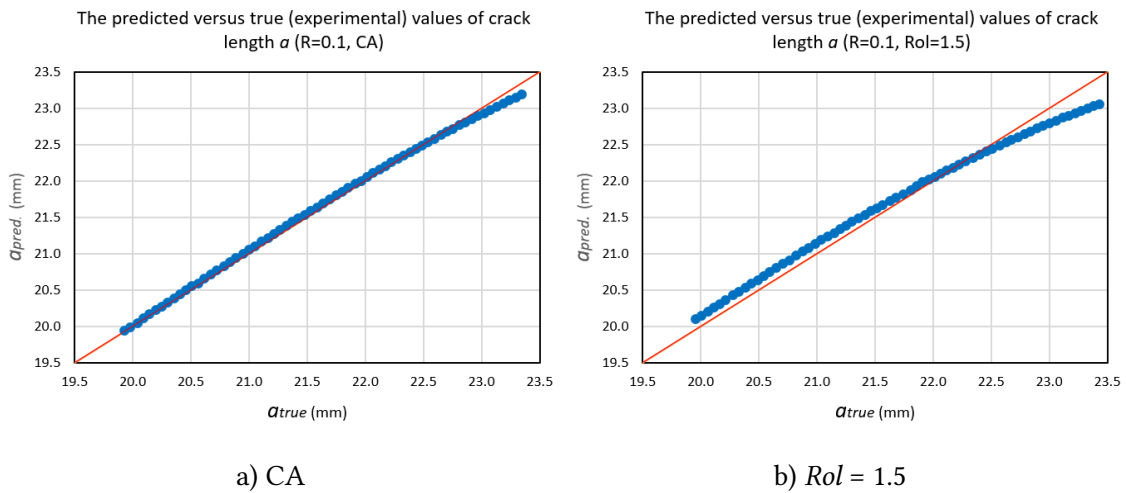

a) CA  b) $Rol = 1.5$

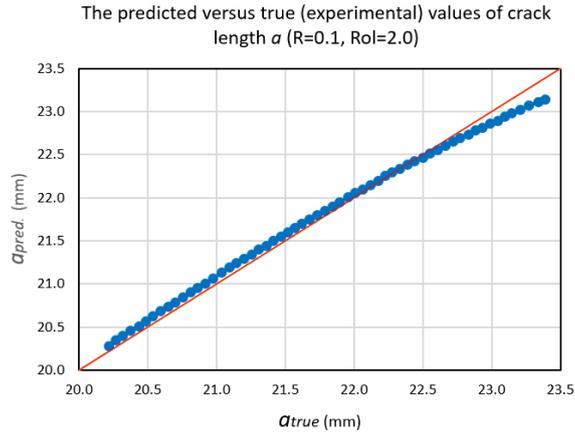

c) $R_{ol} = 2.0$

**Figure 2:** The predicted versus true values of crack length $a$ for stress ratio $R$ = 0.1. a) built for CA; b) built for overload ratio $R_{ol}$ = 1.5; c) built for overload ratio $R_{ol}$ = 2.0

As can be seen from Figure 2, the predicted crack lengths are very close to the experimental values, which is also confirmed by the low value of the MAPE prediction error (Table 1).

**Table 1**
MAPE prediction error for R = 0.1

| Overload ratio $R_{ol}$ | MAPE (%) |
|---|---|
| CA | 0.68 |
| 1.5 | 4.61 |
| 2.0 | 4.59 |

Figure 3 shows the relationship between the experimental values of the crack length $a$ and the predicted values obtained for stress ratio $R$ = 0.3 under constant amplitude and with a single overload with overload factors $R_{ol}$ = 1.5, 2.0.

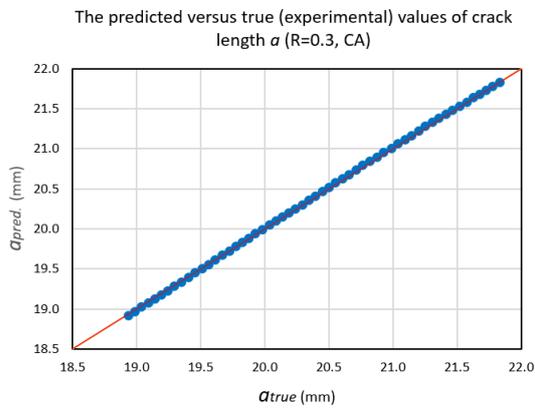
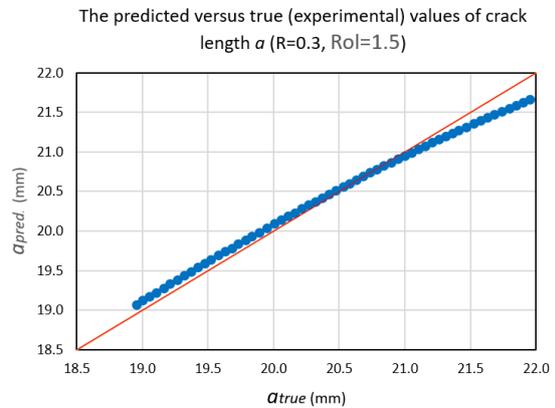

a) CA    b) $R_{ol} = 1.5$

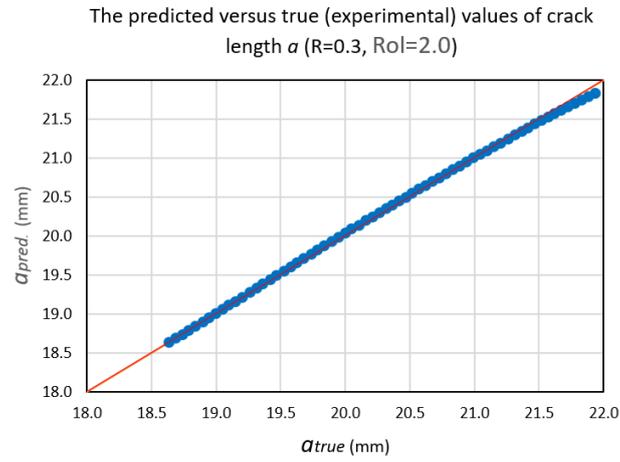

c) $R_{ol} = 2.0$

**Figure 3:** The predicted versus true values of crack length *a* for stress ratio $R = 0.3$. a) built for CA; b) built for overload ratio $R_{ol} = 1.5$; c) built for overload ratio $R_{ol} = 2.0$

With a load factor of $R = 0.3$, the model showed high forecasting accuracy, which is also confirmed by the low value of the MAPE forecasting error (Table 2).

**Table 2**
MAPE prediction error for R = 0.3

| Overload ratio $R_{ol}$ | MAPE (%) |
|---|---|
| CA | 0.07 |
| 1.5 | 3.52 |
| 2.0 | 0.14 |

Figure 4 shows the dependence between the experimental values of the crack length *a* and the predicted values obtained for the load cycle asymmetry factor $R = 0.5$ at CA and $R_{ol} = 1.5, 2.0$.

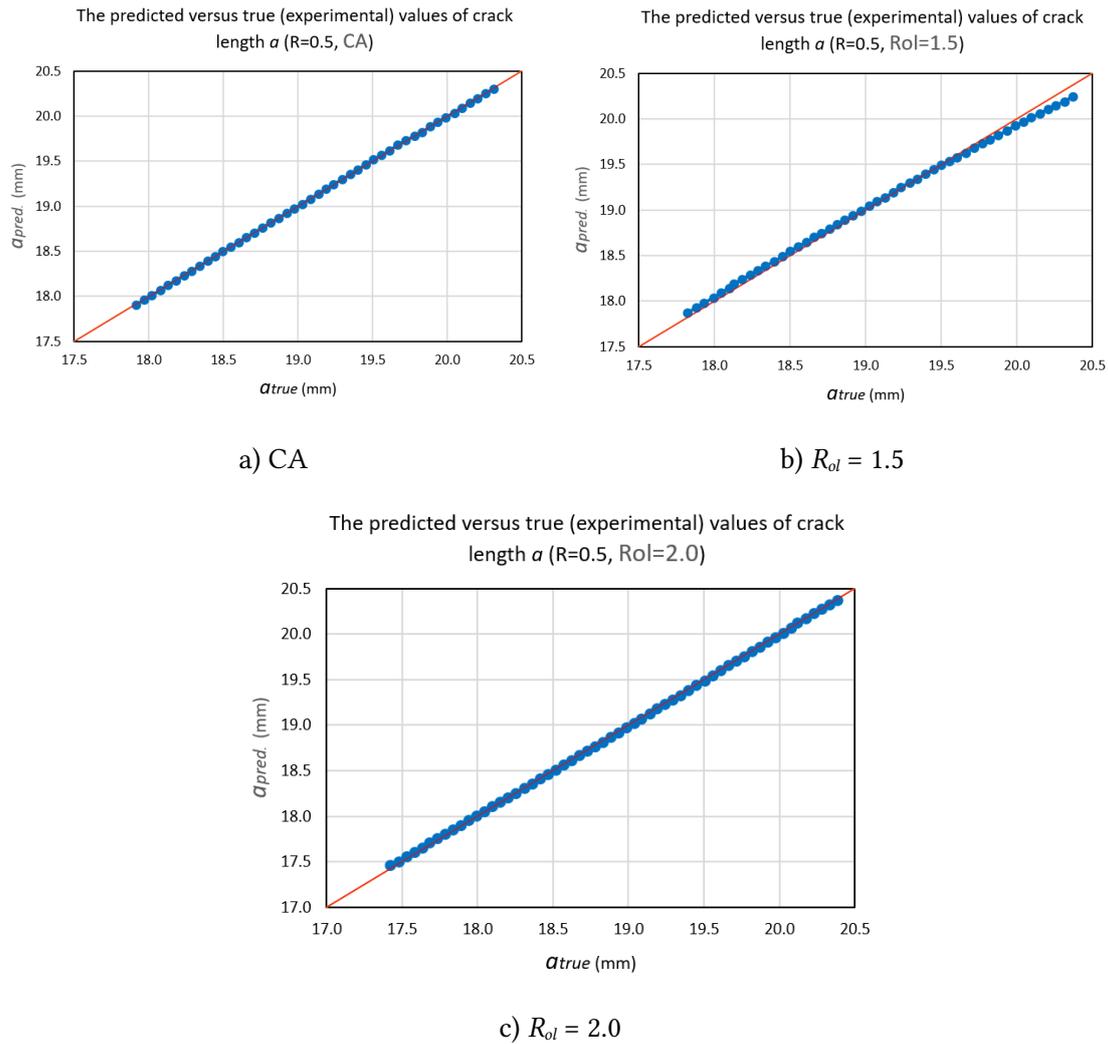

a) CA

b) $R_{ol} = 1.5$

c) $R_{ol} = 2.0$

**Figure 4:** The predicted versus true values of crack length *a* for stress ratio $R$ = 0.5. a) built for CA; b) built for overload ratio $R_{ol}$ = 1.5; c) built for overload ratio $R_{ol}$ = 2.0

At stress ratio $R$ = 0.5, the predicted crack lengths are also very close to the experimental ones, which is confirmed by the low value of the MAPE prediction error (Table 3).

**Table 3**
MAPE prediction error for $R$ = 0.5

| Overload ratio $R_{ol}$ | MAPE (%) |
|---|---|
| CA | 0.02 |
| 1.5 | 0.23 |
| 2.0 | 0.06 |

Figure 5 shows the dependence between the experimental values of the crack length $a$ and the predicted values obtained for a load factor of $R = 0.7$ at CA and $R_{ol} = 1.5, 2.0$.

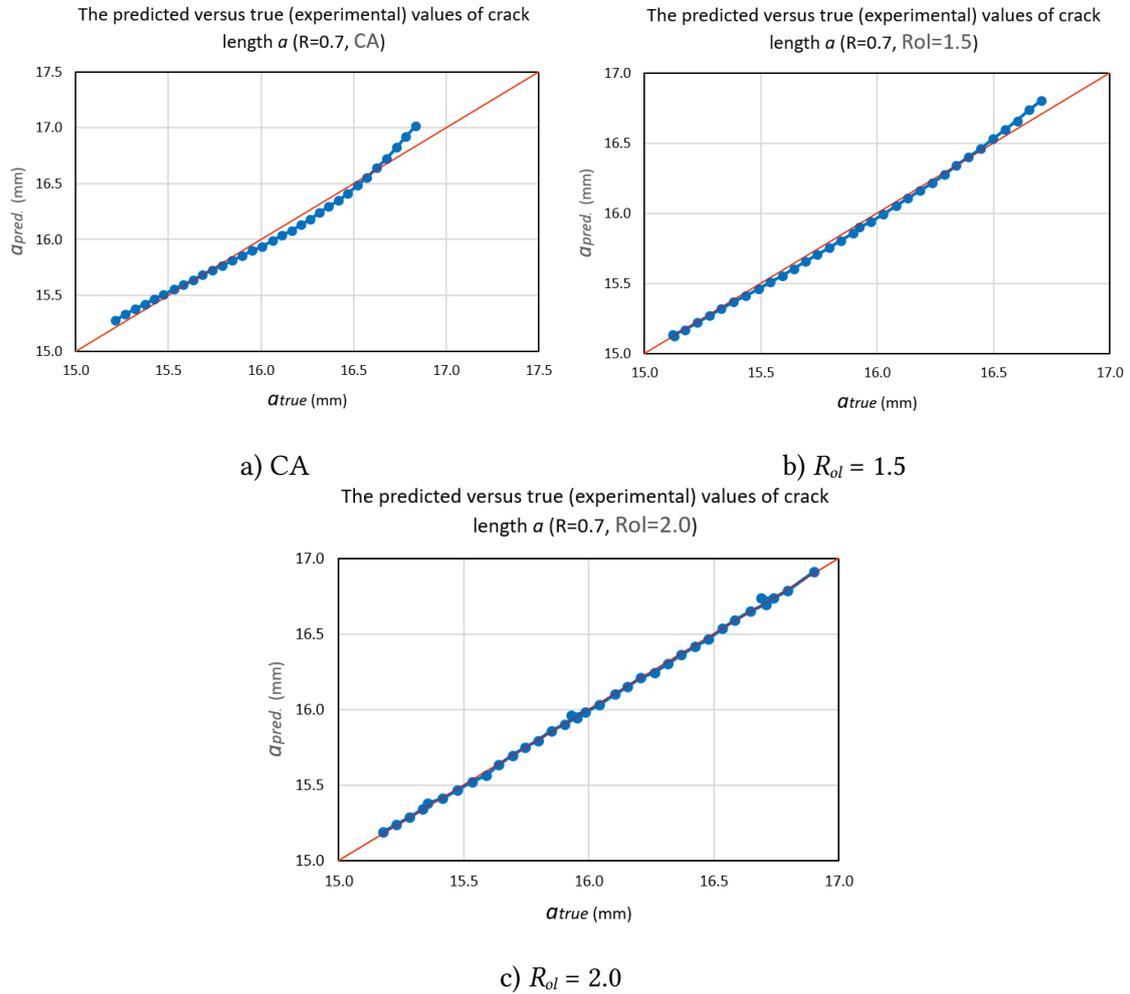

a) CA  
b) $R_{ol} = 1.5$  
c) $R_{ol} = 2.0$

**Figure 5:** The predicted versus true values of crack length $a$ for stress ratio $R = 0.7$. a) built for CA; b) built for overload ratio $R_{ol} = 1.5$; c) built for overload ratio $R_{ol} = 2.0$

Similarly to the previous cases, at a load factor of $R = 0.7$, the predicted crack lengths are quite close to the experimental ones. The values of the MAPE prediction error are given in Table 4.

**Table 4**
MAPE prediction error for $R = 0.7$

| Overload ratio $R_{ol}$ | MAPE (%) |
|---|---|
| CA | 3.49 |
| 1.5 | 0.18 |
| 2.0 | 0.06 |

The obtained results demonstrate the high generalization capability of the model and its effectiveness in reflecting the real fatigue crack growth under cyclic loading.

## 4. Conclusions

The crack length $a$ was predicted as a function of the number of load cycles $N$ for four stress ratio $R$ = 0.1, 0.3, 0.5, and 0.7 at a constant amplitude and overload factors $Rol$ = 1.5, 2.0 by a neural network. The neural network, trained on experimental data, is able to predict the crack length $a$ based on the input parameters, thus providing sufficiently accurate predictions for assessing the fatigue life of QSTE340TM steel.